\documentclass{ws-ijprai}

\begin{document}


%
\catchline{}{}{}{}{}
%

\title{6D Object Pose Estimation Based on 2D Bounding Box}

\author{Jin Liu}

\address{LIESMARS, Wuhan University\\
Wuhan, Hubei, China\\
\email{41038331@qq.com}
}

\author{Sheng He}

\address{LIESMARS, Wuhan Unversity\\
Wuhan, Hubei, China\\
whusheng1996@163.com}

\maketitle


\begin{abstract}
In this paper, we present a simple but powerful method to tackle the problem of estimating the 6D pose of objects from a single RGB image. Our system trains a novel convolutional neural network to regress the unit quaternion, which represents the 3D rotation, from the partial image inside the bounding box returned by 2D detection systems. Then we propose an algorithm we call Bounding Box Equation to efficiently and accurately obtain the 3D translation, using 3D rotation and 2D bounding box. Considering that the quadratic sum of the quaternion’s four elements equals to one, we add a normalization layer to keep the network’s output on the unit sphere and put forward a special loss function for unit quaternion regression. We evaluate our method on the LineMod dataset and experiment shows that our approach outperforms base-line and some state of the art methods.
\end{abstract}


\section{Introduction}

Object detection and localization has always been a hot topic of computer vision. Traditional methods like YOLO[1], SSD[2], and so on, have experienced a tremendous success in 2D domain. However, they can’t achieve accurate semantic understanding of the objective three-dimensional world, for there is no information about the rotation and position between the object and the camera. A pose of a rigid object has 6 degrees of freedom, 3 in translation and 3 in rotation, and its full knowledge is required in many robotic and scene understanding applications[3]. Recently, much attention has been paid to 6D pose estimation. However, estimating the 6D pose of an object is a huge challenge due to various factors, such as the different shapes and visual angles, lighting conditions and occlusions between objects.

Currently, feature-based methods[4,5,6], template-based methods[7,8] and RGB-D methods[9,10,11,12,13] have achieved robust results to some extent. Feature-based methods tackled t33his task by matching feature points between 3D models and images. However, only when there are rich textures on the objects that those methods work. As a result, they are unable to handle texture-less objects[14]. Template-based methods use a rigid template to match different locations in the input image. Such methods are likely to be affected by occlusions. RGB-D methods use depth data as additional information, which simplifies the task. However, active depth sensors are power hungry, which makes 6D objective detection methods for passive RGB images more attractive for mobile and wearable cameras[15]. Besides, acquiring depth data needs additional hardware costs.

Deep learning techniques have recently become mainstream to estimate 6D object pose. In this paper, we propose a generic framework which overcomes the shortcomings of existing methods to estimate 6D object pose. We introduce a brand new method to estimate the 3D rotation \textbf{R} and the 3D translation \textbf{T} from a single RGB image. The key idea of our method is to leverage the results returned by current robust 2D detection systems to recover the 6D pose of objects. Our frame divides the pose estimation task into two main stages. In the first stage, we resize the 2D bounding box returned by a 2D detection system and feed it into a novel convolutional neural network (we call the network Q-Net) to regress the 3D rotation \textbf{R}. In the second stage, our algorithm, Bounding Box Equation, figures out the 3D translation \textbf{T} using \textbf{R} produced in the former stage and the position information of the 2D bounding box on the original image. We choose the unit quaternion \textbf{q} as our 3D rotation representation. The unit quaternion \textbf{q} is a vector of unit length, so the output of the network should be a unit vector, too. Therefore, to enhance the unit quaternion regression, we add a normalization layer to keep the network’s output on the unit sphere, and propose a special loss function, Dot Product Loss, for networks whose outputs require to be unit vectors.

We evaluate our method on the LineMod dataset[17], a benchmark for 6D pose estimation. Experiment shows that on this challenging dataset, our Q-Net and Bounding Box Equation work efficiently and productively, while achieving state-of-the-art results regardless of the complex scenes in the pictures. Additionally, we use our method to detect the 6D pose of common objects in daily life, and the result is also pretty satisfying. In summary, our work has the following advantages and contributions:

1.	Our method doesn’t need any depth data and works on both texture and texture-less images. And it has practical value and can be applied in daily life.

2.	Our method can easily work with current robust 2D detection systems.

3.	We propose a novel convolutional neural network for 3D rotation regression named Q-Net and develop a special loss function, Dot Product Loss, for networks whose outputs require to be unit vectors.

4.	We introduce Bounding Box Equation, a new algorithm to obtain 3D translation using \textbf{R} and 2D bounding box in an efficient and accurate manner.

The remainder of the paper is organized as follows. After the overview of related work, we introduce our approach for 6D object pose estimation. Then we display the experimental results, followed by the final conclusion.

\section{Related Work}

In this section, we review existing techniques designed for 6D pose estimation ranging from traditional approaches to current methods.

\textbf{Traditional approaches.} Feature-based methods and template-based methods are the most traditional techniques in this field. Feature-based methods extract local features described with local descriptors from points of interest in the image, then match them to features on the 3D models to recover the 6D poses[4,5,6,18,19]. For instance, [18] uses SIFT descriptors and clustered images from similar viewpoints into a single model. [19] presents a fast and scalable perception system for object recognition and pose estimation. However, those methods suffer from a common limitation that they require sufficient textures on the objects. To deal with insufficient texture objects, feature learning approaches[17,20] are proposed and outperforms matching approaches. But the basic design of them is time-consuming and multi-stage. In template-based methods[7,8], a rigid template is scanned across the image, and a distance measure is computed to find the best match. These methods can work accurately and quickly, but perform poorly when dealing with clutter and occlusions. 

\textbf{RGB-D approaches.} Depth cameras make RGB-D object pose estimation methods[9,10,11,12,13,21,22,23] prevalent. For example, Brachmann et al. proposed an algorithms suitable for generic objects, both textured and texture-less[9]. Sock et al. proposed a multi-view framework to recognize 6DOF pose of multiple object instances in a crowded scene[21]. Zach et al. developed a dynamic and fast method for RGB-D images. The task can be simplified by using depth images, but acquiring depth data takes extra hardware costs.

\textbf{CNN-based approaches.} In recent years, CNN has become the mainstream to solve 6D pose problems, including camera pose[24, 25] and object pose[26,16,27,28,15,29]. 
Both [24] and [25] train CNNs to directly regress 6D camera pose. Camera pose estimation is much easier for there is no need to detect any object. In [24] the authors use quaternion as rotation representation but omit the spherical constraint, using an unreasonable loss function. [25] addresses the problem by developing a more scientific loss function.

 In [26,16], the authors use CNNs to regress 3D object pose directly, their works focus only on 3D rotation estimation while 3D translation is not included. In [27], SSD detection framework[2] is extended to 6D pose estimation. The authors transform pose detection into two-stage classification tasks, view angle classification and in-plane rotation classification. However, wrong classification in either stage could cause an incorrect pose estimation. In BB8[28], the authors firstly use a segmentation network to localize objects. Then another CNN is used to predict the 2D projections of the 3D bounding box’s corners around the object. The 6D pose is estimated through a PnP algorithm. Finally, a CNN is trained to refine the pose. BB8 achieves high precision but is too time-consuming. Similar to BB8, [15] extends YOLO object detection framework[1] to predict the 2D projections of the corners of the 3D bounding box, then employs PnP algorithm to get the 6D pose. Both [28] and [15] regress too much 2D points, which actually increases the learning difficulty and slows the learning speed. Unlike those mentioned above, Mousavian et al. first use a CNN to regress 3D object orientation, then combines these estimates with geometric constraints provided by a 2D object bounding box to produce a complete 3D bounding box[29]. However, in general, this method needs to solve 4096 linear equations. In special circumstances, such as the KITTI dataset[30], object pitch and roll angles are both zero, there are still 64 equations to be solve, which makes the method computational costly. Our method avoids those problems mentioned above.

\section{Approach}

In this section, we will describe our approach towards 6D object pose estimation.

\subsection{Overall Framework}

The overall pipeline of our approach is shown in Fig. 1. Given a RGB image, we firstly leverage a robust 2D object detection algorithm to locate the object and obtain the 2D bounding box, which is resized to 48*48 and input into Q-Net to regress the unit quaternion \textbf{q}. Then we convert \textbf{q} to rotation matrix \textbf{R} and employ Bounding Box Equation to figure out the 3D translation \textbf{T}. Finally, we project the eight corners of the 3D bounding box onto the image to visualize the 3D rotation and translation.

\begin{figure}[bh]
\centerline{\includegraphics[width=12cm]{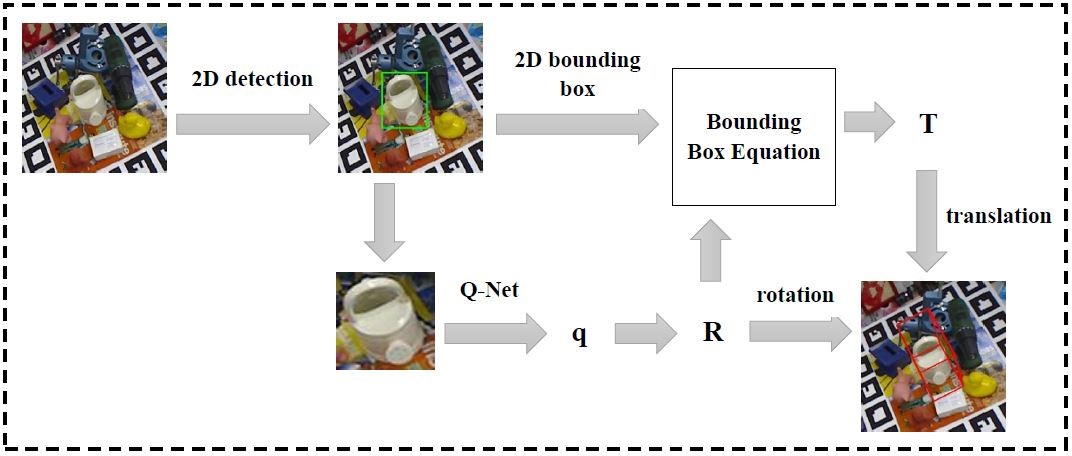}}
\vspace*{8pt}
\caption{Pipeline of our method.}
\end{figure}

\subsection{Q-Net Architecture}

Euler angles, rotation matrix and unit quaternion are three main representations of 3D rotation. Euler angles are easily understandable as they describe three angles of rotation about three axes. However, regressing Euler angles directly can be a hard work due to multiple problems. For example, poses that are very similar visually might be far away in Euler angle space[16]. Rotation matrix is an orthogonal matrix with 3*3 elements and has many special properties. However, regressing it is inappropriate for it is difficult to enforce the orthogonality constraint when learning a 3D rotation representation through back-propagation[25].

Compared to the former two representations, unit quaternion is much more suitable. So to avoid those problems caused by Euler angles and rotation matrix, we choose unit quaternion, \textbf{q} = (q$_{\rm 0}$, q$_{\rm 1}$, q$_{\rm 2}$, q$_{\rm 3}$) as the representation. The only one problem with unit quaternion is that \textbf{q} and \textbf{–q} represent the same rotation. For the loss layer, \textbf{q} and \textbf{–q} will produce the maximum error, which will make the prediction unstable. To address the problem, we constrain all quaternions to one hemisphere. We define \textbf{q$_{\rm +}$} to represent the unit quaternion whose first element is larger than zero. If q$_{\rm 0}$ \textless 0, we change the quaternion to \textbf{q$_{\rm +}$}.

\begin{equation}   
{{q}_{+}}=\left\{\begin{aligned}
  &q,\text{  }{{q}_{0}}>0 \\ 
 -&q,\text{  }{{q}_{0}}\le 0  
\end{aligned} \right.
\end{equation}

To regress unit quaternion, we developed the CNN, Q-Net. The architecture is shown in Fig. 2.

\begin{figure}[bh]
\centerline{\includegraphics[width=12cm]{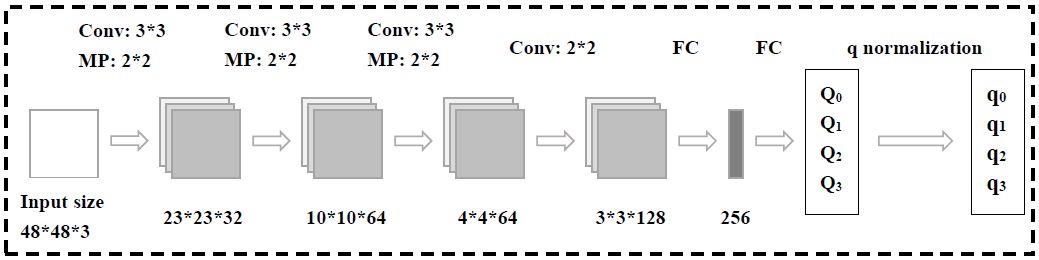}}
\vspace*{8pt}
\caption{The architecture of Q-Net. “Conv” means convolution, “MP” means max pooling, and “FC” means fully connected. The step size of convolution and max pooling is 1 and 2, respectively.}
\end{figure}

Unit quaternion meets the constraint condition:

\begin{equation}   
{{q}_{\text{0}}}^{\text{2}}+{{q}_{\text{1}}}^{\text{2}}+{{q}_{\text{2}}}^{\text{2}}+{{q}_{\text{3}}}^{\text{2}}=1
\end{equation}

However, the fully connected layer doesn’t guarantee that the network output a unit vector. So like [16], we add an additional layer, q normalization layer, to keep the output on the unit sphere. In q normalization layer, the forward-propagation is as follows:

\begin{equation}   
{{q}_{i}}=\frac{{{Q}_{i}}}{\sqrt{Q_{0}^{2}+Q_{1}^{2}+Q_{2}^{2}+Q_{3}^{2}}},i=0,1,2,3
\end{equation}
where Q$_{\rm i}$ is the output of the last fully connected, and q$_{\rm i}$ is the output of q normalization layer.

In [16], the authors have proven that the normalization layer helps improve the accuracy of prediction. Actually, q normalization layer can also accelerate the training of the network. We trained two networks, one with q normalization and the other without that. Apart from the normalization, the two network share the same structure, initialization and samples. We found that with q normalization the loss converged much faster and the training performance is much better, as is shown in Fig. 3.

\begin{figure}[bh]
\centerline{\includegraphics[width=6cm]{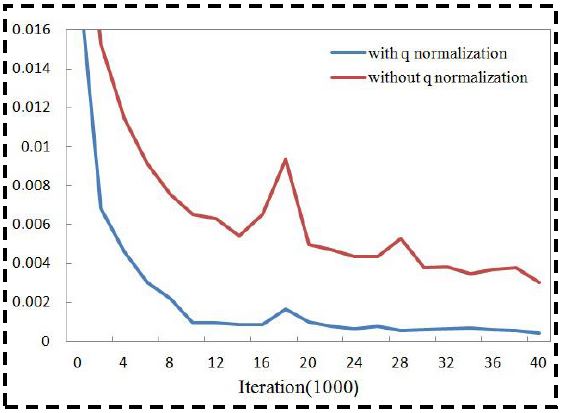}}
\vspace*{8pt}
\caption{The convergence of the loss. The ordinate axis represents the value of the loss, and the abscissa axis represents the iteration.}
\end{figure}

\subsection{Dot Product Loss}

We regard the learning of unit quaternion as a regression problem. Due to the special property of unit quaternion, the network’s output should be a unit vector. To this end, we propose a novel loss function, Dot Product Loss, for networks whose output requires to meet unit constraint. The loss function is defined as follows:

\begin{equation}   
{{L}_{\text{do}t}}=\frac{1}{2}\left( 1-\hat{q}\centerdot q \right)
\end{equation}
where q is the output of network, and q is the ground-truth.

For two unit vectors, their dot product is never larger than one.

\begin{equation}   
\hat{q}\centerdot q=\cos \left( \theta  \right)\le 1
\end{equation}
Only when the two vectors are exactly the same that their dot product equals to one. Under the circumstances, Dot Product Loss reaches the minimum value zero. The neural network reduces the loss through continuous iteration optimization, meanwhile makes the output and ground-truth closer and closer.

When used in regressing unit quaternion, Dot Product Loss is as follows:

\begin{equation}   
{{L}_{dot}}=\frac{1}{2}\left( 1-\sum\limits_{i=0}^{3}{{{{\hat{q}}}_{i}}{{q}_{i}}} \right)
\end{equation}
Both q and q are on the four dimensional sphere. Fig. 4 describes the geometric meaning of Dot Product Loss.

\begin{figure}[bh]
\centerline{\includegraphics[width=6cm]{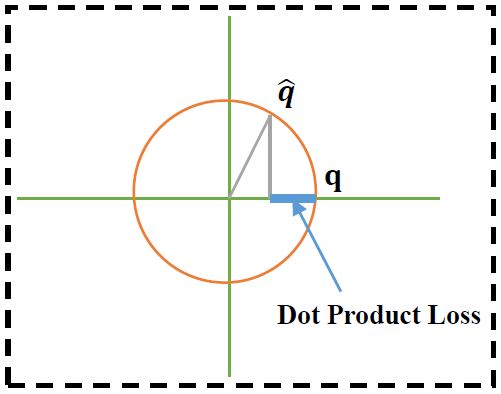}}
\vspace*{8pt}
\caption{The geometric meaning of Dot Product Loss. The red circle refers to the four dimensional sphere.}
\end{figure}
The backward propagation of q normalization layer is as follows:

\begin{equation}   
\frac{\partial {{L}_{\text{dot}}}}{\partial {{Q}_{i}}}=\sum\limits_{j=0}^{3}{\frac{\partial {{L}_{\text{dot}}}}{\partial {{{\hat{q}}}_{j}}}\times \frac{\partial {{{\hat{q}}}_{j}}}{\partial {{Q}_{i}}}}=\sum\limits_{j=0}^{3}{(\text{-}\frac{\text{1}}{\text{2}}{{q}_{j}})\times \frac{\partial {{{\hat{q}}}_{j}}}{\partial {{Q}_{i}}}}=\text{-}\frac{\text{1}}{\text{2}}\sum\limits_{j=0}^{3}{{{q}_{j}}\times \frac{\partial {{{\hat{q}}}_{j}}}{\partial {{Q}_{i}}}}
\end{equation}
where

\begin{equation}   
\left\{ \begin{aligned}
  & \frac{\partial {{{\hat{q}}}_{j}}}{\partial {{Q}_{i}}}=\frac{1}{A}(1-\frac{{{Q}_{i}}^{2}}{{{A}^{2}}})=\frac{1}{A}(1-{{{\hat{q}}}_{i}}^{2})\quad i=j \\ 
 & \frac{\partial {{{\hat{q}}}_{j}}}{\partial {{Q}_{i}}}=\frac{-{{Q}_{i}}{{Q}_{j}}}{{{A}^{3}}}=\frac{-{{{\hat{q}}}_{i}}{{{\hat{q}}}_{j}}}{A}\quad \quad \text{         }i\ne j \\ 
 & A=\sqrt{Q_{0}^{2}+Q_{1}^{2}+Q_{2}^{2}+Q_{3}^{2}} \\ 
\end{aligned} \right.
\end{equation}

\subsection{Bounding Box Equation}

Bounding Box Equation can figure out 3D translation efficiently and accurately, leveraging 3D rotation produced by Q-Net and the 2D bounding box on the original image. The core idea is to use point-to-side correspondence constraint to calculate 3D translation \textbf{T}. The algorithm consists of two step.

\textbf{Step 1 Finding out the four point-to-side correspondence constraints.}

There are n points on the surface of the object, we name them P$_{1}$, P$_{2}$, … , P$_{n}$. Assuming that the origin of the object coordinate frame is in the inside of the object, and the 3D coordinates of those points are X$_{1}$=[x$_{1}$, y$_{1}$, z$_{1}$]$^{T}$, X$_{2}$=[x$_{2}$, y$_{2}$, z$_{2}$]$^{T}$, … , X$_{n}$=[x$_{n}$, y$_{n}$, z$_{n}$]$^{T}$. This step is to find out which four points on the surface are corresponding to the four sides of 2D box, as is shown in Fig. 5. Here we introduce two approach.

\begin{figure}[bh]
\centerline{\includegraphics[width=6cm]{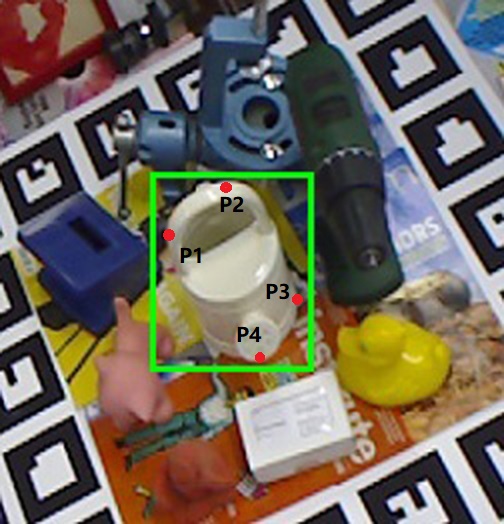}}
\vspace*{8pt}
\caption{The point-to-side correspondence. The surface of the object consists of quantities of points. Here we assume that P$_{1}$, P$_{2}$, P$_{3}$ and P$_{4}$ touch the four sides.}
\end{figure}

\textbf{Approach 1 Indirect comparison.} According to perspective projection,

\begin{equation}   
z\left[ \begin{matrix}
   u  \\
   v  \\
   1  \\
\end{matrix} \right]=KR\left( X-T \right)
\end{equation}
where K is the camera matrix, R is the rotation matrix, u and v are the pixel coordinate, z is the objects’ z-coordinate in the camera coordinate frame and has no impact on the point-to-side correspondence, and it is a positive number. Firstly, we take 2D box’s center (u$_{0}$, v$_{0}$) and the origin of the object coordinate frame, X$_{0}$ = (0, 0, 0)$^{T}$ and a positive number, for example z = 100, into equation (9).

\begin{equation}   
z\left[ \begin{matrix}
   {{u}_{0}}  \\
   {{v}_{0}}  \\
   1  \\
\end{matrix} \right]=KR\left( {{X}_{0}}-T \right)
\end{equation}
We get

\begin{equation}   
{{T}_{0}}=-z{{(KR)}^{-1}}\left[ \begin{matrix}
   {{u}_{0}}  \\
   {{v}_{0}}  \\
   1  \\
\end{matrix} \right]
\end{equation}
T$_{0}$ is not the real 3D translation, but it helps find out the point-to-side correspondence. And we take z, T$_{0}$ into equation (9) to get equation (12).

\begin{equation}   
z\left[ \begin{matrix}
   u  \\
   v  \\
   1  \\
\end{matrix} \right]=KR\left( X-{{T}_{0}} \right)
\end{equation}
Then we take the n points’ 3D coordinates into equation (12) and compute n pairs of pixels coordinates, (u$_{1}$, v$_{1}$), (u$_{2}$, v$_{2}$), … , (u$_{n}$, v$_{n}$). Among those we find the maximum u, v and minimum u, v. The point results in maximum u is corresponding to the right side of the 2D box, we record its index and name it iR. Similarly, the point results in minimum u touches the left side, we record its index and name it iL. The point results in maximum v touches the bottom side, we record its index and name it iB. The point results in minimum v touches the up side, we record its index and name it iT. Up to now we find out which four points are corresponding to the four sides of 2D box.

\textbf{Approach 2 Direct conversion.} For the n points on the surface,

\begin{equation}   
{{z}_{i}}\left[ \begin{matrix}
   {{u}_{i}}  \\
   {{v}_{i}}  \\
   1  \\
\end{matrix} \right]=KR\left( {{X}_{i}}-{{T}_{0}} \right)
\end{equation}
Taking (11) into (13),

\begin{equation}   
{{z}_{i}}\left[ \begin{matrix}
   {{u}_{i}}  \\
   {{v}_{i}}  \\
   1  \\
\end{matrix} \right]=KR\left( {{X}_{i}}+z{{\left( KR \right)}^{-1}}\left[ \begin{matrix}
   {{u}_{0}}  \\
   {{v}_{0}}  \\
   1  \\
\end{matrix} \right] \right)=KR{{X}_{i}}+z\left[ \begin{matrix}
   {{u}_{0}}  \\
   {{v}_{0}}  \\
   1  \\
\end{matrix} \right]
\end{equation}
Establishing a coordinate frame whose origin is at the object’s center, and the three axis are corresponding to the camera's three axis respectively. According to the coordinate transformation principle, in this coordinate frame, the coordinates of point P$_{i}$ is RX$_{i}$ = [$\Delta$X$_{i}$, $\Delta$Y$_{i}$, $\Delta$Z$_{i}$]. And we get

\begin{equation}   
KR{{X}_{i}}=\left[ \begin{matrix}
   {{f}_{x}}\Delta {{X}_{i}}+{{c}_{x}}\Delta {{Z}_{i}}  \\
   {{f}_{y}}\Delta {{Y}_{i}}+{{c}_{y}}\Delta {{Z}_{i}}  \\
   \Delta {{Z}_{i}}  \\
\end{matrix} \right]=\left[ \begin{matrix}
   {{P}_{xi}}  \\
   {{P}_{yi}}  \\
   \Delta {{Z}_{i}}  \\
\end{matrix} \right]
\end{equation}
where f$_{x}$, f$_{y}$, c$_{x}$, c$_{y}$ are the elements of camera matrix. z is the objects’ z-coordinate in the camera coordinate frame an is much larger than $\Delta$Z$_{i}$. so,

\begin{equation}   
\frac{\Delta {{Z}_{i}}}{z}\approx \text{0}
\end{equation}
From (14), (15) and (16) we get

\begin{equation}   
\left\{ \begin{aligned}
  & {{u}_{i}}=\frac{{{f}_{x}}\Delta {{X}_{i}}+{{c}_{x}}\Delta {{Z}_{i}}+z{{u}_{0}}}{\Delta {{Z}_{i}}+z}=\frac{\frac{{{f}_{x}}\Delta {{X}_{i}}}{z}+\frac{{{c}_{x}}\Delta {{Z}_{i}}}{z}+{{u}_{0}}}{\frac{\Delta {{Z}_{i}}}{z}+\text{1}}\approx \frac{{{f}_{x}}\Delta {{X}_{i}}}{z}+{{u}_{0}} \\ 
 & {{v}_{i}}=\frac{{{f}_{y}}\Delta {{Y}_{i}}+{{c}_{y}}\Delta {{Z}_{i}}+z{{v}_{0}}}{\Delta {{Z}_{i}}+z}=\frac{\frac{{{f}_{y}}\Delta {{Y}_{i}}}{z}+\frac{{{c}_{y}}\Delta {{Z}_{i}}}{z}+{{v}_{0}}}{\frac{\Delta {{Z}_{i}}}{z}+\text{1}}\approx \frac{{{f}_{y}}\Delta {{Y}_{i}}}{z}+{{v}_{0}} \\ 
\end{aligned} \right.
\end{equation}
From (17) we can see that the values of u$_{i}$ and v$_{i}$ are determined by $\Delta$X$_{i}$ and $\Delta$Y$_{i}$ if we assign a positive number to z. (15) can simplify to

\begin{equation}   
R{{X}_{i}}=\left[ \begin{matrix}
   \Delta {{X}_{i}}  \\
   \Delta {{Y}_{i}}  \\
   \Delta {{Z}_{i}}  \\
\end{matrix} \right]
\end{equation}
We take n points' 3D coordinates into (18) and calculate n pair of [$\Delta$X, $\Delta$Y, $\Delta$Z], through which we can infer which four points result in the maximum and minimum u, v. And we find out the four points corresponding to the four sides of 2D bounding box.

\textbf{Step 2 Figuring out 3D translation}

In the former step, we find out the four points and name their indexes iL, iR, iT and iB. According to collinearity equation,

\begin{equation}   
\left\{ \begin{aligned}
  & {{u}_{\text{K}i}}=\frac{{{u}_{i}}\text{-}{{\text{c}}_{\text{x}}}}{f}=\frac{{{r}_{11}}({{x}_{i}}-{{\text{t}}_{x}})+{{r}_{12}}({{y}_{i}}-{{\text{t}}_{y}})+{{r}_{13}}({{z}_{i}}-{{\text{t}}_{z}})}{{{r}_{31}}({{x}_{i}}-{{\text{t}}_{x}})+{{r}_{32}}({{y}_{i}}-{{\text{t}}_{y}})+{{r}_{33}}({{z}_{i}}-{{\text{t}}_{z}})} \\ 
 & {{\text{v}}_{\text{K}i}}=\frac{{{v}_{i}}\text{-}{{\text{c}}_{\text{y}}}}{f}=\frac{{{r}_{21}}({{x}_{i}}-{{\text{t}}_{x}})+{{r}_{22}}({{y}_{i}}-{{\text{t}}_{y}})+{{r}_{23}}({{z}_{i}}-{{\text{t}}_{z}})}{{{r}_{31}}({{x}_{i}}-{{\text{t}}_{x}})+{{r}_{32}}({{y}_{i}}-{{\text{t}}_{y}})+{{r}_{33}}({{z}_{i}}-{{\text{t}}_{z}})} \\ 
\end{aligned} \right.
\end{equation}
where (t$_{x}$, t$_{y}$, t$_{z}$) is the elements of T and unknown, T is the camera’s 3D coordinate in the object coordinate frame. (x$_{i}$, y$_{i}$, z$_{i}$) is the 3D coordinate. We linearize (19) and get

\begin{equation}   
\left\{ \begin{aligned}
  & ({{u}_{\text{K}i}}{{r}_{31}}-{{r}_{11}}){{t}_{x}}+({{u}_{\text{K}i}}{{r}_{32}}-{{r}_{12}}){{t}_{y}}+({{u}_{\text{K}i}}{{r}_{33}}\text{-}{{r}_{13}}){{t}_{z}}= \\ 
 & ({{u}_{\text{K}i}}{{r}_{31}}-{{r}_{11}}){{x}_{i}}+({{u}_{\text{K}i}}{{r}_{32}}-{{r}_{12}}){{y}_{i}}+({{u}_{\text{K}i}}{{r}_{33}}\text{-}{{r}_{13}}){{z}_{i}} \\ 
 & ({{v}_{\text{K}i}}{{r}_{31}}-{{r}_{21}}){{t}_{x}}+({{v}_{\text{K}i}}{{r}_{32}}-{{r}_{22}}){{t}_{y}}+({{v}_{\text{K}i}}{{r}_{33}}\text{-}{{r}_{23}}){{t}_{z}}= \\ 
 & ({{v}_{\text{K}i}}{{r}_{31}}-{{r}_{21}}){{x}_{i}}+({{v}_{\text{K}i}}{{r}_{32}}-{{r}_{22}}){{y}_{i}}+({{v}_{\text{K}i}}{{r}_{33}}\text{-}{{r}_{23}}){{z}_{i}} \\ 
\end{aligned} \right.
\end{equation}
According to the four point-to-side correspondence constraints, we get

\begin{equation}   
\left\{ \begin{aligned}
  & ({{u}_{\text{K}iL}}{{r}_{31}}-{{r}_{11}}){{t}_{x}}+({{u}_{\text{K}iL}}{{r}_{32}}-{{r}_{12}}){{t}_{y}}+({{u}_{\text{K}iL}}{{r}_{33}}\text{-}{{r}_{13}}){{t}_{z}}= \\ 
 & ({{u}_{\text{K}iL}}{{r}_{31}}-{{r}_{11}}){{x}_{iL}}+({{u}_{\text{K}iL}}{{r}_{32}}-{{r}_{12}}){{y}_{iL}}+({{u}_{\text{K}iL}}{{r}_{33}}\text{-}{{r}_{13}}){{z}_{iL}} \\ 
 & ({{u}_{\text{K}iR}}{{r}_{31}}-{{r}_{11}}){{t}_{x}}+({{u}_{\text{K}iR}}{{r}_{32}}-{{r}_{12}}){{t}_{y}}+({{u}_{\text{K}iR}}{{r}_{33}}\text{-}{{r}_{13}}){{t}_{z}}= \\ 
 & ({{u}_{\text{K}iR}}{{r}_{31}}-{{r}_{11}}){{x}_{iR}}+({{u}_{\text{K}iR}}{{r}_{32}}-{{r}_{12}}){{y}_{iR}}+({{u}_{\text{K}iR}}{{r}_{33}}\text{-}{{r}_{13}}){{z}_{iR}} \\ 
 & ({{v}_{\text{K}i\text{T}}}{{r}_{31}}-{{r}_{21}}){{t}_{x}}+({{v}_{\text{K}i\text{T}}}{{r}_{32}}-{{r}_{22}}){{t}_{y}}+({{v}_{\text{K}iT}}{{r}_{33}}\text{-}{{r}_{23}}){{t}_{z}}= \\ 
 & ({{v}_{\text{K}iT}}{{r}_{31}}-{{r}_{21}}){{x}_{i\text{T}}}+({{u}_{\text{K}iT}}{{r}_{32}}-{{r}_{\text{2}2}}){{y}_{i\text{T}}}+({{u}_{\text{K}i\text{T}}}{{r}_{33}}\text{-}{{r}_{\text{2}3}}){{z}_{i\text{T}}} \\ 
 & ({{v}_{\text{K}iB}}{{r}_{31}}-{{r}_{21}}){{t}_{x}}+({{u}_{\text{K}iB}}{{r}_{32}}-{{r}_{22}}){{t}_{y}}+({{v}_{\text{K}iB}}{{r}_{33}}\text{-}{{r}_{23}}){{t}_{z}}= \\ 
 & ({{v}_{\text{K}iB}}{{r}_{31}}-{{r}_{21}}){{x}_{i\text{B}}}+({{u}_{\text{K}iB}}{{r}_{32}}-{{r}_{\text{2}2}}){{y}_{i\text{B}}}+({{u}_{\text{K}i\text{B}}}{{r}_{33}}\text{-}{{r}_{\text{2}3}}){{z}_{i\text{B}}} \\ 
\end{aligned} \right.
\end{equation}
(21) can be written in matrix form as follows:

\begin{equation}   
AT={{X}_{box}}
\end{equation}
where

\begin{equation}   
A=\left[ \begin{matrix}
   {{u}_{\text{K}iL}}{{r}_{31}}-{{r}_{11}} & {{u}_{\text{K}iL}}{{r}_{32}}-{{r}_{12}} & {{u}_{\text{K}iL}}{{r}_{33}}\text{-}{{r}_{13}}  \\
   {{u}_{\text{K}iR}}{{r}_{31}}-{{r}_{11}} & {{u}_{\text{K}iR}}{{r}_{32}}-{{r}_{12}} & {{u}_{\text{K}iR}}{{r}_{33}}\text{-}{{r}_{13}}  \\
   {{v}_{\text{K}i\text{T}}}{{r}_{31}}-{{r}_{21}} & {{v}_{\text{K}i\text{T}}}{{r}_{32}}-{{r}_{22}} & {{v}_{\text{K}iT}}{{r}_{33}}\text{-}{{r}_{23}}  \\
   {{v}_{\text{K}iB}}{{r}_{31}}-{{r}_{21}} & {{u}_{\text{K}iB}}{{r}_{32}}-{{r}_{22}} & {{v}_{\text{K}iB}}{{r}_{33}}\text{-}{{r}_{23}}  \\
\end{matrix} \right]=\left[ \begin{matrix}
   {{b}_{Left}}  \\
   {{b}_{Right}}  \\
   {{b}_{Top}}  \\
   {{b}_{Bottom}}  \\
\end{matrix} \right]
\end{equation}
We call equation (22) Bounding Box Equation, the matrix A Bounding Box Matrix, and the four row vectors Side Vector. In (23),

\begin{equation}   
\left\{ \begin{aligned}
  & {{u}_{\text{K}i\text{L}}}=\frac{{{\text{x}}_{\text{L}}}\text{-}{{\text{c}}_{\text{x}}}}{{{f}_{x}}} \\ 
 & {{u}_{\text{K}i\text{R}}}=\frac{{{\text{x}}_{\text{R}}}\text{-}{{\text{c}}_{\text{x}}}}{{{f}_{y}}} \\ 
 & {{\text{v}}_{\text{K}iT}}=\frac{{{y}_{T}}\text{-}{{\text{c}}_{\text{y}}}}{{{f}_{x}}} \\ 
 & {{\text{v}}_{\text{K}iB}}=\frac{{{y}_{B}}\text{-}{{\text{c}}_{\text{y}}}}{{{f}_{y}}} \\ 
\end{aligned} \right.
\end{equation}
where x$_{L}$, y$_{L}$, x$_{R}$, y$_{R}$ are the pixel coordinates of the 2D bounding box’s borders. In Bounding Box Equation,

\begin{equation}   
{{X}_{box}}=\left[ \begin{matrix}
   ({{u}_{\text{K}iL}}{{r}_{31}}-{{r}_{11}}){{x}_{iL}}+({{u}_{\text{K}iL}}{{r}_{32}}-{{r}_{12}}){{y}_{iL}}+({{u}_{\text{K}iL}}{{r}_{33}}\text{-}{{r}_{13}}){{z}_{iL}}  \\
   ({{u}_{\text{K}iR}}{{r}_{31}}-{{r}_{11}}){{x}_{iR}}+({{u}_{\text{K}iR}}{{r}_{32}}-{{r}_{12}}){{y}_{iR}}+({{u}_{\text{K}iR}}{{r}_{33}}\text{-}{{r}_{13}}){{z}_{iR}}  \\
   ({{v}_{\text{K}iT}}{{r}_{31}}-{{r}_{21}}){{x}_{i\text{T}}}+({{u}_{\text{K}iT}}{{r}_{32}}-{{r}_{\text{2}2}}){{y}_{i\text{T}}}+({{u}_{\text{K}i\text{T}}}{{r}_{33}}\text{-}{{r}_{\text{2}3}}){{z}_{i\text{T}}}  \\
   ({{v}_{\text{K}iB}}{{r}_{31}}-{{r}_{21}}){{x}_{i\text{B}}}+({{u}_{\text{K}iB}}{{r}_{32}}-{{r}_{\text{2}2}}){{y}_{i\text{B}}}+({{u}_{\text{K}i\text{B}}}{{r}_{33}}\text{-}{{r}_{\text{2}3}}){{z}_{i\text{B}}}  \\
\end{matrix} \right]=\left[ \begin{matrix}
   {{b}_{Left}}\bullet {{X}_{iL}}  \\
   {{b}_{Right}}\bullet {{X}_{iR}}  \\
   {{b}_{Top}}\bullet {{X}_{iT}}  \\
   {{b}_{Bottom}}\bullet {{X}_{iB}}  \\
\end{matrix} \right]
\end{equation}
We call the four row vectors Bounding Box Vector. The norm of Side Vector is as follows:

\begin{equation}   
\left\{ \begin{aligned}
  & ||{{b}_{Left}}|{{|}^{2}}=||\{{{u}_{\text{K}iL}}{{r}_{31}}-{{r}_{11}},{{u}_{\text{K}iL}}{{r}_{32}}-{{r}_{12}},{{u}_{\text{K}iL}}{{r}_{33}}\text{-}{{r}_{13}}\}|{{|}^{2}}=u_{_{\text{K}iL}}^{2}+1 \\ 
 & ||{{b}_{Right}}|{{|}^{2}}=||\{{{u}_{\text{K}iR}}{{r}_{31}}-{{r}_{11}},{{u}_{\text{K}iR}}{{r}_{32}}-{{r}_{12}},{{u}_{\text{K}iR}}{{r}_{33}}\text{-}{{r}_{13}}\}|{{|}^{2}}=u_{_{\text{K}iR}}^{2}+1 \\ 
 & ||{{b}_{Top}}|{{|}^{2}}=||\{{{v}_{\text{K}iT}}{{r}_{31}}-{{r}_{\text{2}1}},{{v}_{\text{K}iT}}{{r}_{32}}-{{r}_{\text{2}2}},{{v}_{\text{K}iT}}{{r}_{33}}\text{-}{{r}_{\text{2}3}}\}|{{|}^{2}}=v_{_{\text{K}iT}}^{2}+1 \\ 
 & ||{{b}_{Bottom}}|{{|}^{2}}=||\{{{v}_{\text{K}iB}}{{r}_{31}}-{{r}_{\text{2}1}},{{v}_{\text{K}iB}}{{r}_{32}}-{{r}_{\text{2}2}},{{v}_{\text{K}iB}}{{r}_{33}}\text{-}{{r}_{\text{2}3}}\}|{{|}^{2}}=v_{_{\text{K}iB}}^{2}+1 \\ 
\end{aligned} \right.
\end{equation}
To solve Bounding Box Equation, we use least square method.

\begin{equation}   
T={{\left( A{{A}^{T}} \right)}^{-1}}{{A}^{T}}{{X}_{box}}
\end{equation}
Here, the T we get is the camera’s 3D coordinate in the object coordinate frame. 3D translation \textbf{T} is the object’s 3D coordinate in the camera coordinate. \textbf{T} can be obtained by the follow equation:

\begin{equation}   
\mathbf{T}=-RT
\end{equation}

\section{Experiments}

In this section, we present and discuss the results of our method.

\subsection{Implementation Details}

We test our approach both on LineMod dataset[17] and daily objects. For there is only one object of interest on each image, we choose MTCNN[31] as the 2D detection system to obtain the 2D bounding box. Then use Q-Net and Bounding Box Equation to estimate the 6D pose.

We train Q-Net by SGD with momentum, using a mini-batch size of 24 images and variable learning rates, “step” is the learning rate policy. The step size is set 10$^{4}$ and gamma is set 0.8. Other parameters are set as follows: momentum 0.9, weight decay 4*10$^{-3}$, and base learning rate 10$^{-4}$. Xavier initialization is chosen for weights and zero-initialize biases. Dropout is included where used behind the full connection layer. All models are trained and tested with Caffe.

\subsection{6D Pose Estimation Results}

Fig. 6 and Fig. 7 respectively show the results of 6D pose estimation on LineMod and daily objects. LineMod provides a point cloud for each object, so we use all the 3D coordinates of the point cloud in Bounding Box Equation. Actually, there is no need to use so much points. While testing on the daily objects, no point cloud is provided, we use eight corners to estimate the 3D translation.

\begin{figure}[bh]
\centerline{\includegraphics[width=7cm]{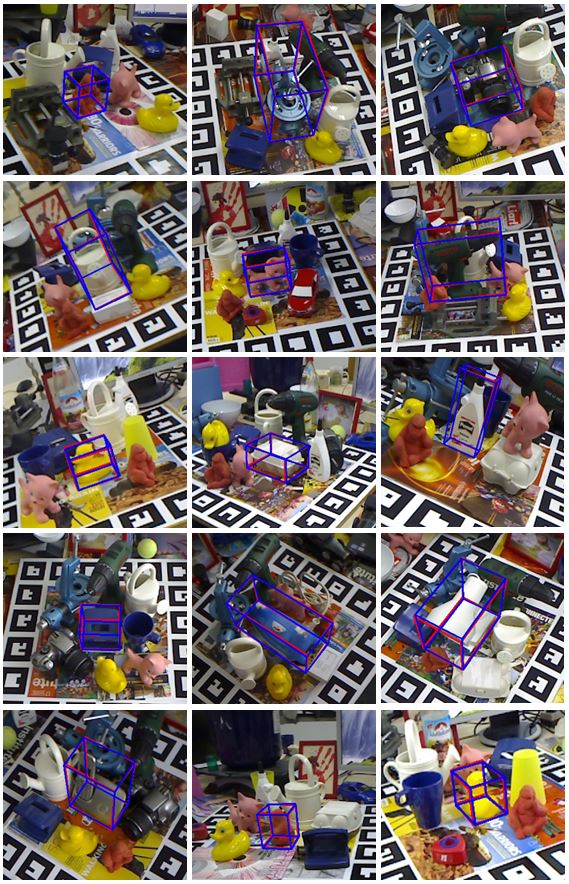}}
\vspace*{8pt}
\caption{Pose estimation results of our method on LineMod dataset[17]. The blue 3D bounding boxes refer to the ground-truth, and the red refer to the estimation results.}
\end{figure}

To our knowledge, there is no benchmark for daily objects 6D pose estimation, due to which present methods are confined to the laboratory environment and haven’t been widely applied. In our experiments, all the training data of daily objects are annotated by ourselves.

\begin{figure}[bh]
\centerline{\includegraphics[width=7cm]{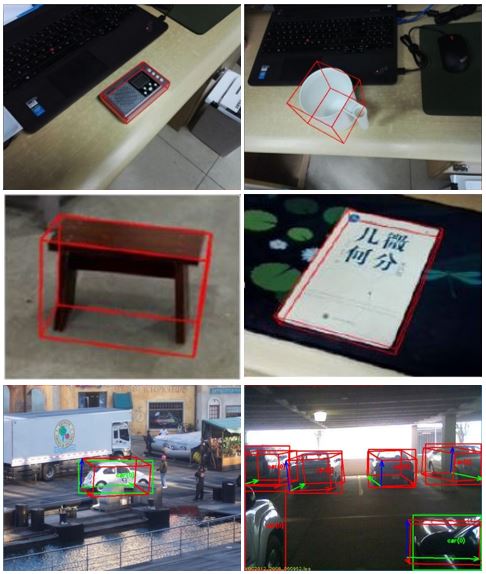}}
\vspace*{8pt}
\caption{Pose estimation results of our method on daily objects. The first line shows a radio and a cup, the second shows a stool and a book. These pictures are taken by ourselves. The third line shows the results on cars, the pictures are taken from Pascal VOC dataset[32].}
\end{figure}

\subsection{State Of The Art Comparisons}

To evaluate our method and compare it with some state of the art methods, we use two evaluation metrics.

\textbf{Average Euler angel error.} This is a metric appropriate for 3D rotation estimation accuracy evaluation. It is used in [16] and matches the one used in [33]. Both [16] and [33] train CNNs to regress the 3D rotation. Average Euler angle error is the average of the absolute difference in angle (in degrees) between the estimated 3D rotation and the ground truth, measured regarding the three principal axes[16]. Since we use unit quaternion representation, we transform it into Euler angle to perform the comparisons. Table 1 shows the comparisons with [16] and [33].

\begin{table}[th]
\tbl{State of the art comparisons of our method against the one of Wohlhart et al.[33] and Doumanoglou et al.[16] in the LineMod dataset[17]. The former two don’t provide results for eggbox and glue, hence for the sake of comparison the average is taken over the first 11 objects.}
{\begin{tabular}{@{}cccc@{}} \toprule
Object    & [   33   ] & [   16   ] & \textbf{   ours   }  \\   \botrule   
ape\hphantom{00} & \hphantom{}15.0 & \hphantom{}11.8 & \textbf{5.3} \\
benchviseblue\hphantom{00} & \hphantom{}15.5 & \hphantom{}13.2 & \textbf{6.1} \\
camera\hphantom{00} & \hphantom{}12.0 & \hphantom{}10.1 & \textbf{4.8} \\
can\hphantom{00} & \hphantom{}15.5 & \hphantom{}12.3 & \textbf{4.9} \\ 
cat\hphantom{00} & \hphantom{}14.0 & \hphantom{}10.4 & \textbf{4.6} \\
driller\hphantom{00} & \hphantom{}17.8 & \hphantom{}13.2 & \textbf{5.9} \\
duck\hphantom{00} & \hphantom{}13.9 & \hphantom{}10.9 & \textbf{5.0} \\
holepuncher\hphantom{00} & \hphantom{}13.2 & \hphantom{}11.4 & \textbf{5.5} \\ 
iron\hphantom{00} & \hphantom{}11.4 & \hphantom{}10.2 & \textbf{6.2} \\
lamp\hphantom{00} & \hphantom{}13.3 & \hphantom{}11.1 & \textbf{4.2} \\
phone\hphantom{00} & \hphantom{}18.2 & \hphantom{}11.7 & \textbf{4.8} \\
\textbf{average}\hphantom{00} & \hphantom{}14.53 & \hphantom{}11.48 & \textbf{5.21} \\ 
eggbox\hphantom{00} & \hphantom{}- & \hphantom{}- & \textbf{5.1} \\
glue\hphantom{00} & \hphantom{}- & \hphantom{}- & \textbf{5.2} \\ \botrule
\end{tabular}}
\end{table}

\textbf{5cm 5$^{o}$.} This is a metric appropriate for 6D pose estimation accuracy evaluation. As in [34], we use the percentage of correctly predicted poses for each object. With this metric, a pose is considered correct if the translational and rotational error[3] are below 5cm and 5$^{o}$ respectively. The translational error (e$_{TE}$) and rotational error (e$_{RE}$) are as follows:

\begin{equation}   
{{e}_{TE}}\left( \hat{t},\bar{t} \right)={{\left\| \bar{t}-\hat{t} \right\|}_{2}}
\end{equation}

\begin{equation}   
{{e}_{RE}}\left( \hat{R},\bar{R} \right)=\arccos \left( \left( Tr\left( \hat{R}{{{\bar{R}}}^{-1}} \right)-1 \right)/2 \right)
\end{equation}
where  and   are the estimated pose,  and  are the ground truth pose. The error e$_{RE}$ is given by the angle from the axis-angle representation of rotation[35]. In [34] the authors introduce an approach to estimate 6D pose from a single RGB image. Our method relies on RGB images, too. Table 2 shows the comparisons with [34].

\begin{table}[th]
\tbl{State of the art comparisons of our method against the one of Wohlhart et al.[33] and Doumanoglou et al.[16] in the LineMod dataset[17]. The former two don’t provide results for eggbox and glue, hence for the sake of comparison the average is taken over the first 11 objects.}
{\begin{tabular}{@{}cccc@{}} \toprule
Object    & [   34   ] &  \textbf{   ours   }  \\   \botrule   
ape\hphantom{00} & \hphantom{}34.4 &  \textbf{55.0} \\
benchviseblue\hphantom{00} & \hphantom{}40.6 & \textbf{44.0} \\
camera\hphantom{00} & \hphantom{}30.5 & \textbf{62.5} \\
can\hphantom{00} & \hphantom{}48.4  & \textbf{64.8} \\ 
cat\hphantom{00} & \hphantom{}34.6 & \textbf{67.7} \\
driller\hphantom{00} & \hphantom{}54.5 & \textbf{50.0} \\
duck\hphantom{00} & \hphantom{}22.0 & \textbf{58.4} \\
eggbox\hphantom{00} & \hphantom{}57.1 & \textbf{60.0} \\
glue\hphantom{00} & \hphantom{}23.6 & \textbf{59.6} \\ 
holepuncher\hphantom{00} & \hphantom{}47.3 & \textbf{55.6} \\ 
iron\hphantom{00} & \hphantom{}58.74 & \textbf{46.7} \\
lamp\hphantom{00} & \hphantom{}49.3 & \textbf{70.1} \\
phone\hphantom{00} & \hphantom{}26.8 & \textbf{60.5} \\
\textbf{average}\hphantom{00} & \hphantom{}40.60 & \textbf{58.07} \\ \botrule
\end{tabular}}
\end{table}

\subsection{Computation Times}

On LineMod dataset, our implementation takes 16ms for 2D detection (MTCNN), 2ms for 3D rotation prediction, and 1ms for Bounding Box Equation, on a GeForce GTX 1080 GPU.

\section{Conclusion}

In this work, we propose a method that extracts 6D pose from a single RGB image based on 2D bounding box. This method inherits the previous 2D object detection algorithm very well. On this basis, we train Q-Net on Dot Product Loss to regress the unit quaternion and use Bounding Box Equation to obtain 3D translation. Experiments show that the method is feasible, efficient and practical. Up to now, our method is limited in single type objects pose estimation. As a future work, we would like to further improve the performance and focus on how to extend this method in order to tackle multi type objects pose estimation. Moreover, we plan to extend the dataset to daily objects.


\vspace*{-0.01in}
\noindent
\rule{12.6cm}{.1mm}

\end{document}